\documentclass[sigconf]{acmart}
\AtBeginDocument{%
  \providecommand\BibTeX{{%
    \normalfont B\kern-0.5em{\scshape i\kern-0.25em b}\kern-0.8em\TeX}}}

\copyrightyear{2021}
\acmYear{2021}
\setcopyright{acmcopyright}\acmConference[HotMobile '21]{The 22nd International Workshop on Mobile Computing Systems and Applications}{February 24--26, 2021}{Virtual, United Kingdom}
\acmBooktitle{The 22nd International Workshop on Mobile Computing Systems and Applications (HotMobile '21), February 24--26, 2021, Virtual, United Kingdom}
\acmPrice{15.00}
\acmDOI{10.1145/3446382.3448362}
\acmISBN{978-1-4503-8323-3/21/02}

\usepackage{multirow}
\usepackage{algorithm2e}
\usepackage{subcaption}
\usepackage{booktabs}
\usepackage{caption}
\usepackage{subcaption}
\usepackage{multirow}
\usepackage{array}
\usepackage{pgfplots}
\usepackage{soul}

\SetKwBlock{Initialization}{initialization}{}
\SetKwBlock{ForwardPass}{forward pass}{}
\SetKwBlock{BackwardPass}{backward pass}{}
\SetKwInOut{Input}{input}
\SetKwInOut{Output}{output}

\newcommand\blfootnote[1]{%
  \begingroup
  \renewcommand\thefootnote{}\footnote{#1}%
  \addtocounter{footnote}{-1}%
  \endgroup
}

\begin{document}

\title{SplitEasy: A Practical Approach for Training ML models on Mobile Devices}

\author{Kamalesh Palanisamy}
\email{kamalesh800@gmail.com}
\affiliation{%
  \institution{NIT Trichy}
  \country{}
}

\author{Vivek Khimani}
\email{vck29@drexel.edu}
\affiliation{%
  \institution{Drexel University}
  \country{}
}

\author{Moin Hussain Moti}
\email{mhmoti@cse.ust.hk}
\affiliation{%
  \institution{HKUST}
  \country{}
}

\author{Dimitris Chatzopoulos}
\email{dcab@cse.ust.hk}
\affiliation{%
  \institution{HKUST}
  \country{}
}
\renewcommand{\shortauthors}{Palanisamy, Khimani, Moti, Chatzopoulos}

\begin{abstract}

Modern mobile devices, although resourceful, cannot train state-of-the-art machine learning models without the assistance of servers, which require access to, potentially, privacy-sensitive user data. Split learning has recently emerged as a promising technique for training complex deep learning (DL) models on low-powered mobile devices. The core idea behind this technique is to train the sensitive layers of a DL model on mobile devices while offloading the computationally intensive layers to a server. Although a lot of works have already explored the effectiveness of split learning in simulated settings, a usable toolkit for this purpose does not exist. In this work, we highlight the theoretical and technical challenges that need to be resolved to develop a functional framework that trains ML models in mobile devices without transferring raw data to a server. Focusing on these challenges, we propose \textit{SplitEasy}, a framework for training ML models on mobile devices using split learning. Using the abstraction provided by \textit{SplitEasy}, developers can run various DL models under split learning setting by making minimal modifications.
We provide a detailed explanation of \textit{SplitEasy} and perform experiments with six state-of-the-art neural networks. We demonstrate how \textit{SplitEasy} can train models that cannot be trained solely by a mobile device while incurring nearly constant time per data sample. \blfootnote{For the code of \textit{SplitEasy} please visit the following link and contact Kamalesh Palanisamy for any concerns: \href{https://github.com/kamalesh0406/SplitEasy}{https://github.com/kamalesh0406/SplitEasy}}
\end{abstract}




\maketitle

\section{Introduction}

Deep learning (DL) is a widely adopted technique in mobile applications. The basic idea behind applications that use DL is to train and embed neural network (NN) models on users' data to learn their usage patterns. Thus, enabling the models to predict user actions and assist in various tasks such as text auto-completion, email tagging, and content recommendation. However, these applications do not train models on the device because of the limited hardware capabilities. Instead, they upload user data to the cloud and train the models on powerful servers. Notably, this exposure to the outside environment compromises users' privacy.

One solution is to employ transfer learning (TL) techniques~\cite{pan2010tl}, which involve using pre-trained models on public data as the starting point of on-device training. Using TL, one can also use light machine learning frameworks like Squeezenet~\cite{Iandola2017SqueezeNet} for reduced model size. Pre-trained models require less effort than training models from scratch, but the process is still inefficient compared to cloud training, mainly due to hardware limitations. Federated learning (FL)~\cite{konecny2016fedl} is another machine learning paradigm that proposes training several instances of a model on separate devices and then aggregating all these instances into a unifying model. However, this paradigm also suffers from similar issues because the user device still needs to perform training on some part of the data to capture its specific usage pattern. Hence, it is impossible to avoid inefficient on-device training while safeguarding users' privacy using current learning frameworks.

In this work, we develop \textit{SplitEasy}, a novel framework, that solves this problem by splitting the training procedure into three parts, the first and the last parts are handled by the mobile device, and a server handles the middle part. More specifically, as shown in Figure~\ref{fig:overview}, for a multilayered network, we execute the first few layers on the device and then send the output for further training to a server; the server then propagates the received data through many in-between hidden layers before returning them to the device; finally, the device executes the last few layers of the network and computes the training loss. The loss similarly propagates in the opposite direction (\textit{backpropagates}). Thus, we only exchange the output of intermediate layers that conveys no meta information about the input data or labels. For example, in the case of image classification, the images are input only to the mobile-side and then transformed to floating points (irrelevant to an outside entity) after the first few layers. Next, the output returned from the server is propagated through the last few layers on the mobile side before comparing that to the labels stored only on the mobile side. Overall, outside entities know neither the input data nor what the model is trying to learn in our framework, thus preserving the user's privacy. Moreover, since almost the whole network executes on the server-side, it is equivalent to training on the cloud.

\textit{SplitEasy} is flexible and can work with simple vanilla NN architectures to complex CNN-based architectures like Densenet~\cite{huang2017densely} and ResNet~\cite{he2016deep}. Even the devices need not be mobile devices. In fact, any computer with minimal computational capabilities will work with our framework. Similarly, any computer with high enough computational power can act as a server. Users can set different split limits for their devices depending on the type of network connection in usage, neural network architecture employed, and the device and server specifications. To this extent, we perform a thorough empirical analysis of our framework in varying settings. 

In summary, the major contributions of this work are: \textit{(i)} an in-depth explanation of our privacy-aware and robust split learning framework for mobile devices, \textit{(ii)} an empirical analysis in varying network settings to show the versatility of \textit{SplitEasy}, \textit{(iii)} the open-source code of \textit{SplitEasy} for the benefit of the research community.

\section{Split Learning Challenges}~\label{sec:splitLearning}
Split learning techniques partition neural networks into multiple sections where each section is located in a different machine. Each section iteratively performs its tasks and passes on the output to the next or the previous section depending on the direction of propagation. Altogether, these disjoint sections must emulate one complete neural network. Existing theoretical frameworks and software tools are not designed to complement these settings. Therefore, in order to realize split learning, we propose some theoretical modifications to the backpropagation algorithm and discuss how to tackle the system challenges in this section.

\subsection{Theoretical Challenges}\label{sec:backprob}
Deep learning frameworks such as PyTorch \cite{PyTorch} and Tensorflow \cite{Tensorflow} use automatic differentiation (AD)~\cite{AutoDiff} to dynamically compute the gradients for performing optimization tasks such as backpropagation~\cite{backprop}. AD is a collection of techniques employed to calculate the derivatives of a function. A typical DL framework represents the variables and operations as a directed graph, with the final node being a custom loss function. AD computes the gradients of intermediate layers by backtracing through the graph that begins at the node, which represents the loss function. 

\smallskip\noindent\textbf{Automatic Differentiation.} However, when a split learning architecture splits a network between multiple models, graphs get disconnected, and so AD techniques can no longer be used to directly compute the gradients for layers present on models apart from the last model where the loss is computed. 
In addition, the technical differences between frameworks used for mobile and server implementation result in additional costs associated with data manipulation, transformation, and parsing. Therefore, we propose a novel, platform-independent approach for updating the model purely based on the gradients acquired from the previous split, instead of relying on the overall loss, which is computed on the last model.
The rest of this section provides a theoretical background of our approach.

We will focus on a double split architecture where a model is split twice, that is, into three separate models. The first and the third model reside independently on the mobile device, while the second model is hosted on a server.
$f_A(x)$ denotes the function representing the layers of the first model, $f_B(x)$ represents the layers of the second model, and $f_C(x)$ the layers of the third model.

\begin{figure}[t]
  \centering
  \includegraphics[width=\columnwidth]{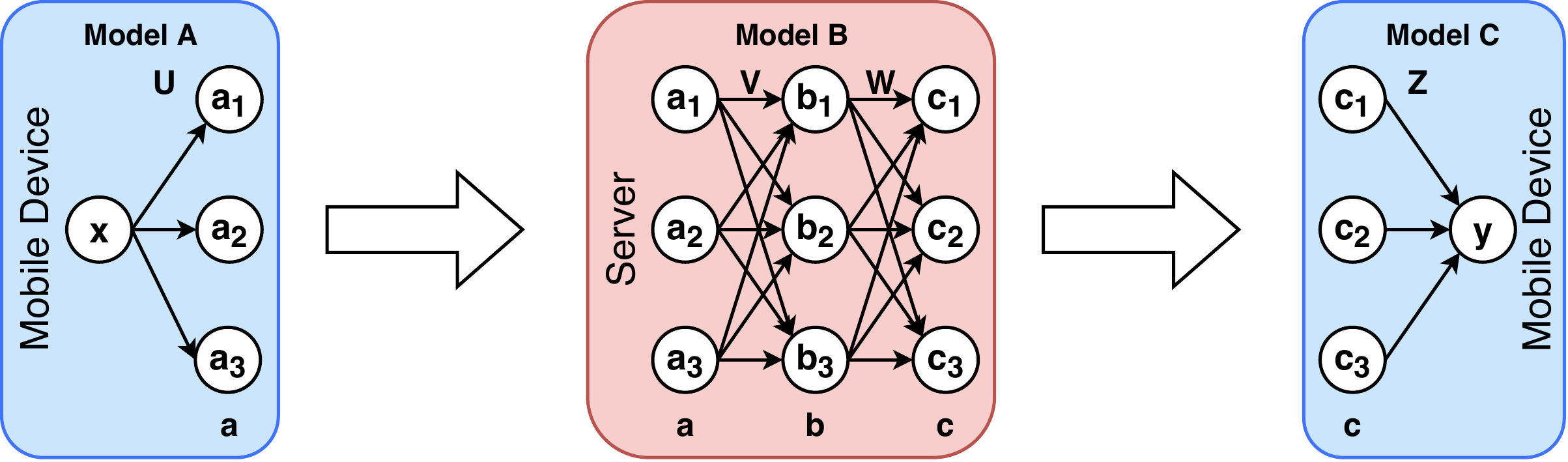}
  \caption{Overview of split learning with a simple NN.}
  \label{fig:overview}
\end{figure}

For the sake of simplicity we consider a model with four layers and assume that $f_A$ contains the first layer, $f_B$ contains the next two layers, and $f_C$ contains the last layer, as shown in Figure~\ref{fig:overview}. 
Hence the output $\hat{y}$ of the network with input $x$ is:
\begin{equation}
\hat{y} = f_C(f_B(f_A(x)))
\end{equation}
Let $y_t$ be the target for the network. The loss $L$ calculated using the objective function $J$ can be represented as:
\begin{equation}
L = J( y_t, \hat{y}) 
\end{equation}    

Following the common practice, we consider as an objective function $J$ the mean squared error (MSE) in a regression problem and the cross-entropy loss (CE) in a classification problem. The forward pass equations for each of the layers can be represented as: 
\begin{align*}
\mathbf{a} &= g(U^T.\mathbf{x} + U_b) \hspace{1cm}\mathbf{b} = h(V^T.\mathbf{a} + V_b) \\
\mathbf{c} &= i(W^T.\mathbf{b} + W_b) \hspace{1cm} \mathbf{\hat{y}} = j(Z^T.\mathbf{c} + Z_b) 
\end{align*}
where $\{a, g(x), U, U_b\}, \{b, h(x), V, V_b\}, \{c, i(x), W, W_b\},\{\hat{y},\\ j(x), Z, Z_b\}$ represent the outputs, activation functions, weights and biases of each of the layers respectively. The update equation for weight $Z$ on the mobile device at time step $t$ is:
\begin{equation}
    Z_t = Z_{t-1} - \eta  \frac{\partial L}{\partial Z_{t-1}}
\end{equation}
where $Z_{t-1}$ represents the weights at time step $t-1$ and $\eta$ is the learning rate. The learning rate $\eta$ remains the same for Models A, B and C. The update equation for $W$ at time step $t$ on the server is:
\begin{eqnarray}
    W_t = W_{t-1} - \eta  \frac{\partial L}{\partial W_{t-1}}
    \label{eq:winc} 
 \text{, where } \frac{\partial L}{\partial W_{t-1}} = \frac{\partial L}{\partial c}  \frac{\partial c}{\partial W_{t-1}}
\end{eqnarray}

$\frac{\partial L}{\partial c}$ can be sent from the mobile device to the server, but the weights cannot be updated without AD. Therefore, we propose using auxiliary labels such that when models update their values using these labels, the result is equivalent to the scenario when there is no split.

We first send the gradients $\frac{\partial L}{\partial c}$ from the mobile device to the server. For Model B, in the server, we define the auxiliary loss as~$L_B$:
\begin{equation}
    L_B = \frac{1}{2}\sum (\hat{c} - c)^2 \text{, where } \hat{c} = c + \frac{\partial L}{\partial c}
\end{equation}
We use residual sum squared error (RSSE) to update the weights in $W$ instead of MSE as it leads to better gradients. Since there is no dependency on $N$ in Equation~\ref{eq:winc}, using MSE will reduce the gradient updates in $W$ by a factor of $N$. The updated equation for $W$ under the new loss becomes:
\begin{eqnarray}
    W_t &=& W_{t-1} - \eta  \frac{\partial L_B}{\partial W_{t-1}}\label{eq:winb} \\
        \text{where, }\hspace{0.6cm}\frac{\partial L_B}{\partial W_{t-1}} &=& \frac{\partial L_B}{\partial c}  \frac{\partial c}{\partial W_{t-1}}  = \frac{1}{2}  2  (\hat{c} - c)  \frac{\partial c}{\partial W_{t-1}}
\end{eqnarray}
Substituting $\hat{c}$ above we get,
\begin{equation}
    \begin{split}
            \frac{\partial L_B}{\partial W_{t-1}} = (c + \frac{\partial L}{\partial c} - c)  \frac{\partial c}{\partial W_{t-1}} = \frac{\partial L}{\partial c}  \frac{\partial c}{\partial W_{t-1}}
            \label{eq:clossderiv}
    \end{split}
\end{equation}
By substituting Equation~\ref{eq:clossderiv} in Equation~\ref{eq:winb} we get the same equation as Equation~\ref{eq:winc}. Thus, we can update the weights of the model in the server although there is no direct connection between the models on the mobile device and the server.
The main advantage of this implementation is that it works on both the server and the mobile device since it just requires modifying the targets and the loss.



\subsection{System Challenges}

In addition to the theoretical challenges, split learning is prone to a multitude of system challenges when deploying such techniques in mobile devices. Whenever split learning techniques are implemented in client-server architectures, it is challenging to maintain a persistent socket connection between the client and the server while having a low communication cost. Furthermore, as split learning allocates a major part of devices' hardware for on-device training and transmits high volumes of data, it has a notable impact on traditional metrics of mobile computing such as energy consumption and mobile data usage. The main reason behind not considering the issues of resource utilization and energy consumption is the popular assumption of running the learning tasks when the device is idle and connected to the Internet via Wifi (e.g. when the user is sleeping at night). We design SplitEasy using frameworks that provide flexibility of accessing and updating the model weights, as discussed in Section~\ref{sec:overview}. Before introducing the design of SplitEasy we list two additional system challenges that are not considered in this work and are part of our future work.

\smallskip\noindent\textbf{Data Privacy.} 
SL was considered to be a privacy-preserving technique because it does not require any exchange of raw data.
However, recent advances in deep learning have demonstrated that gradients or trained models shared with the server are vulnerable to information leakage and attacks that can potentially violate data privacy~\cite{154,155,156}. Unless such privacy concerns are resolved, split learning can be considered as a privacy-aware architecture, instead of a privacy-preserving one.

\smallskip\noindent\textbf{Mobile Operating Systems.} Every learning task that is executed in a mobile device can be either part of a conventional mobile application or run as a background process. Considering that a learning task may need minutes or even hours when executed in a mobile device, it is only reasonable if they run in the background without disrupting the regular workflow of the mobile users. However, as the mobile devices are required to make multiple API calls to the server for every iteration, running the training process in the background is a challenging problem. As the background processes are managed by the mobile operating system, they are not allowed to make the API calls as and when required for training. In fact, to the best of our knowledge, the current state of modern mobile operating system only allows for a few calls to background processes per day, which would make the training process highly inefficient.

\begin{figure}[t]
  \centering
  \includegraphics[width=1\linewidth]{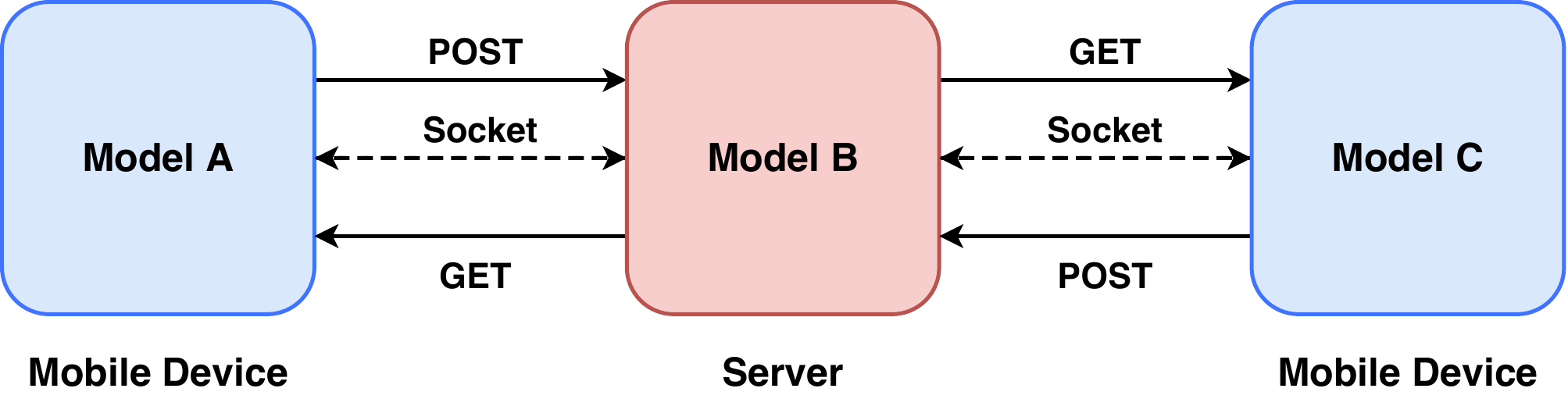}
  \caption{Interactions between models in \textit{SplitEasy}.\vspace{-0.3cm}}
  \label{fig:implementation}
\end{figure}

\section{Overview of SplitEasy}\label{sec:overview}

\textit{SplitEasy} can function between any number of devices, but for this paper, we will stick with a basic client-server architecture as shown in Figure~\ref{fig:implementation}. The client is a mobile device that acts as both the starting and the ending point. That is, at any point of time, there are two separate instances running in parallel on the mobile device. We denote the start point as model A and the endpoint as model C. We use a single stand-alone server as model B.

A reliable connection between the models is crucial to the working of \textit{SplitEasy}. Therefore, the first step is to establish a persistent socket connection between the mobile device and the server.
Next, the data is pre-processed on the mobile device. Data can be input in batches. Let $B$ denote the batch size of the data. We input the data to model A. After propagating through the first few layers of the network, model A sends the generated output to Model B through a POST request.
The size of data sent is equal to $B \cdot R \cdot C \cdot F$, where $R \times C$ are the dimensions of the output features and $F$ denotes the number of filters.
Model B propagates the received data through its layers (that is, most of the hidden layers) and broadcasts a message notifying the listeners of the task completion.
Model C listening on its end, receives the message and sends a GET request to model B to get its output values. The size of the data received is equal to $B \cdot E$, where $E$ denotes the embedding size. These values are subsequently propagated through the layers of Model C.

Since model C is the endpoint of the network, it holds all the labels for computing the loss of the network. For model C, we simply use these labels to compute the loss and gradients to backpropagate through its layers. Note that the output from model B acts as the input layer for model C; we compute the gradients of this layer as well and then issue a POST request from model C to model B to send these gradients. Model B uses these gradients to generate its labels. Once the labels are obtained, we follow the standard procedure to compute the loss and update the model's weights. Next, similar to model C, model B computes the gradient of its input layer (that is, the output received from model A) and notifies model A to collect it. The size of these gradients is equal to $B \cdot R \cdot C \cdot F$. Model A then collects these gradients using a GET request and uses them to generate its labels. After that, we follow the standard procedure for backpropagation.


\section{Performance Evaluation}\label{sec:experiments}
We implemented \emph{SplitEasy} in such a way to be compatible with existing technical support. We list below the implementation details (Section~\ref{sec:setup}), and the used datasets (Section~\ref{sec:data}), and discuss the conducted experiments to highlight its performance (Section~\ref{sec:exps}).

\subsection{Setup}\label{sec:setup}
We evaluated the performance of \emph{SplitEasy} using an iPhone for the mobile side and three difference servers for the server side. 

\smallskip\noindent\textbf{Mobile Framework.} Android does not provide any official support for training deep learning models, but only inferencing with pre-trained models. Therefore, we chose iOS and iPhone XR with 3GB RAM as our mobile platform. On iOS, we had four deep learning libraries to choose from, namely, \emph{Tensorflow}, \emph{CoreML}, \emph{Metal Performance Shaders} (MPS) and \emph{LibTorch}. There is no official Tensorflow API for iOS yet, but it provides compatible API(s) in the form of \emph{Tensorflow.js}. We chose TensorFlow.js (with TensorFlow React Native) for implementing our framework as we found the following limitations in the other libraries:
\begin{enumerate}
    \item CoreML and MPS: It is critical for our framework to be able to access all the gradients and define custom models. Both CoreML and MPS are very difficult to customize and access all the variables.
    \item LibTorch: This is Pytorch's C++ library, which could be used for the implementation on iOS with an Objective-C wrapper. But the library only supports CPU computations for mobile devices.
\end{enumerate}

After considering all these factors, we selected TensorFlow.js since it provides access to weights and gradients and uses a \emph{WebGL} backend for GPU computations. 



\smallskip\noindent\textbf{Server Framework.} We use three different servers in our experiments, a home server with an NVIDIA 1660 GPU, an Amazon Web Services (AWS) instance with an NVIDIA K80 GPU, and a university server with an NVIDIA RTX 2080 GPU. In all three servers, we installed Python Flask\footnote{Flask: \url{https://flask.palletsprojects.com/en/1.1.x/}} and Pytorch\footnote{Pytorch: \url{https://pytorch.org/}} to implement the functionality of the server, as presented in the Section~\ref{sec:overview}.

Although it is very common to use the number of FLOPS executed to compare split learning with other architectures~\cite{gupta2018distributed,vepakomma2018split}, this metric is not fair as it does not take into account the time spent on communication with other models.  
It is not surprising that the number of FLOPS executed on the mobile device is far less in split learning as only a small part of the network is executed on it. We, therefore, use runtime as the comparison metric in our experiments. We use image datasets for all our experiments as it is the most popular use case, but the framework can be easily extended to support more cases.
Moreover, since mobile users have a highly variable number of images in their devices, we report all statistics with respect to a single image, that is, time taken to train the model on a single image.

\begin{figure}[t]
  \centering
  \includegraphics[width=\columnwidth]{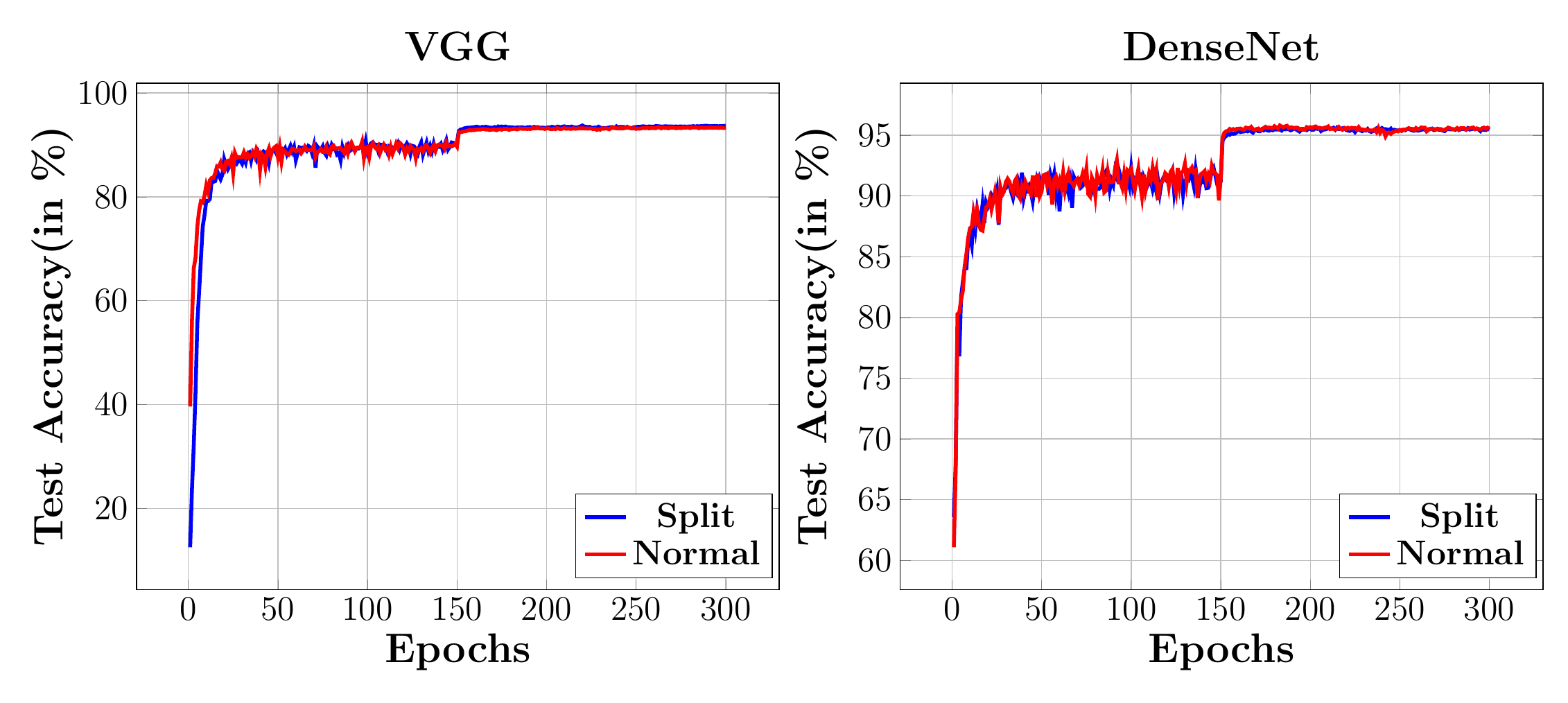}
  \caption{Accuracy comparison between traditional stochastic gradient decent and \textit{SplitEasy}.}
  \label{fig:proof_of_concept}
\end{figure}

\begin{figure*}[t]
  \centering
\begin{subfigure}{0.59\columnwidth}
  \centering
  \includegraphics[width=\columnwidth]{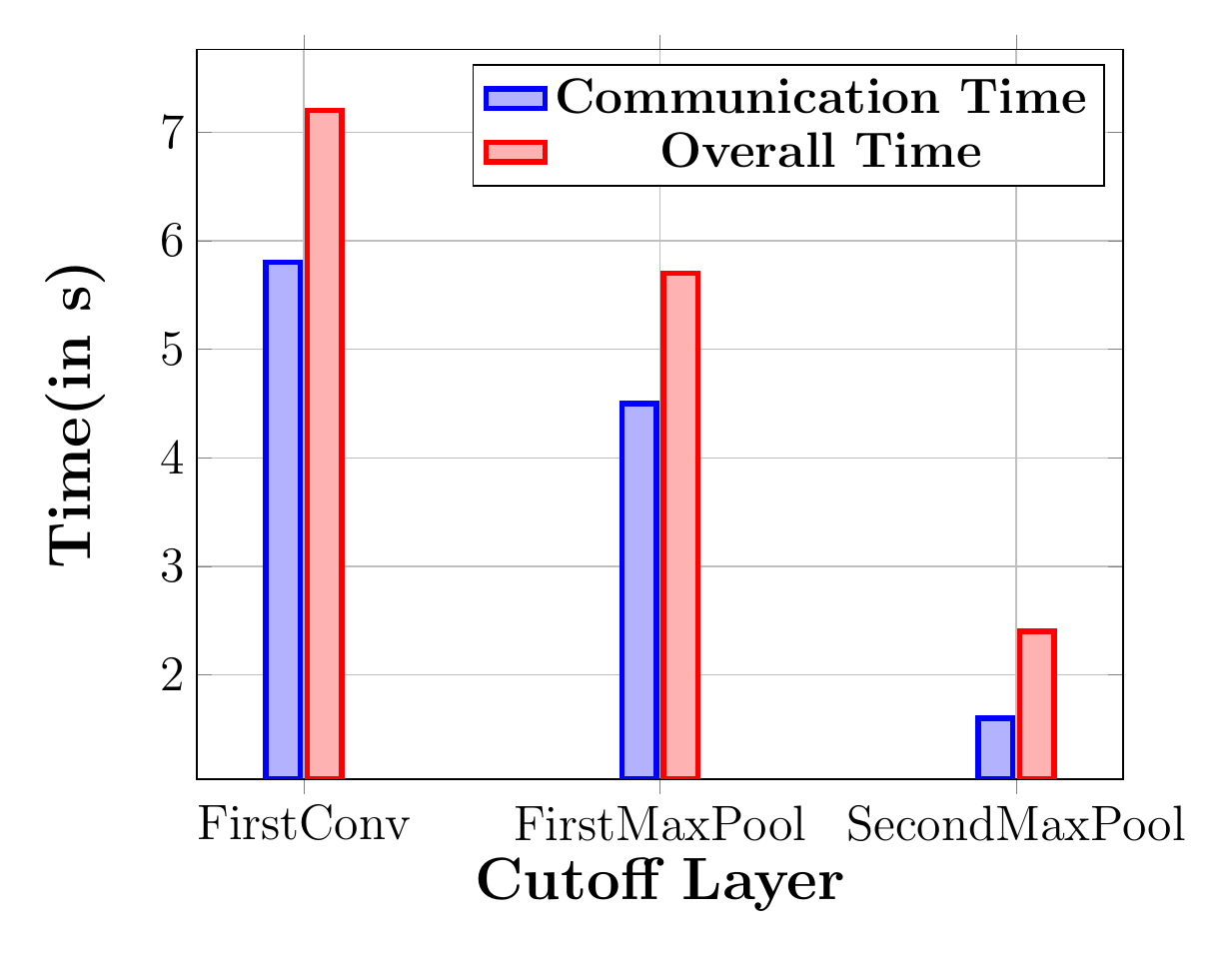}
  \caption{Different cutoffs.}
  \label{fig:cutoff}
\end{subfigure}
\begin{subfigure}{0.59\columnwidth}
  \centering
   \includegraphics[width=\columnwidth]{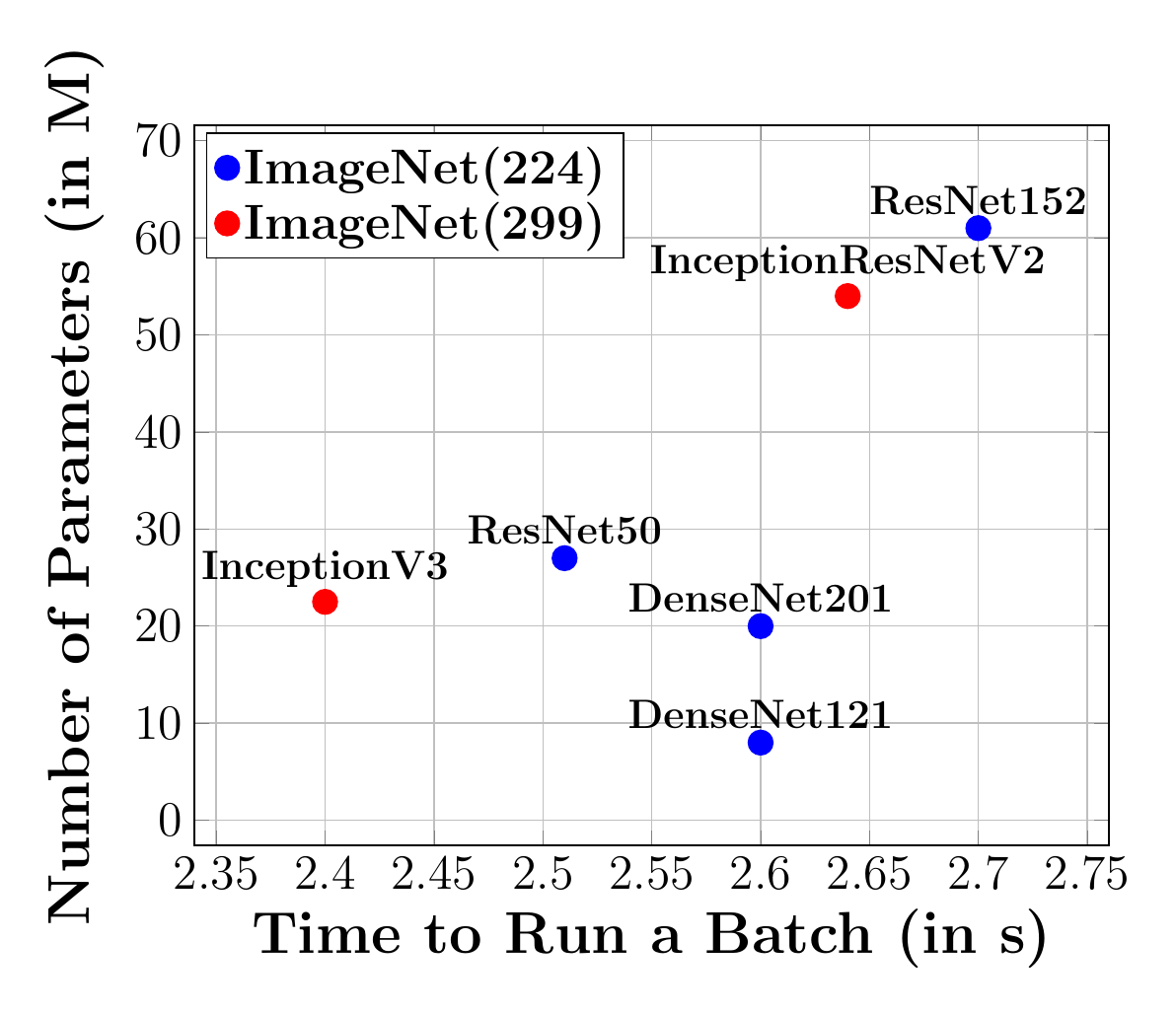}
  \caption{Different models.}
  \label{fig:scatter}
\end{subfigure}
\begin{subfigure}{0.59\columnwidth}
  \centering
\includegraphics[width=\columnwidth]{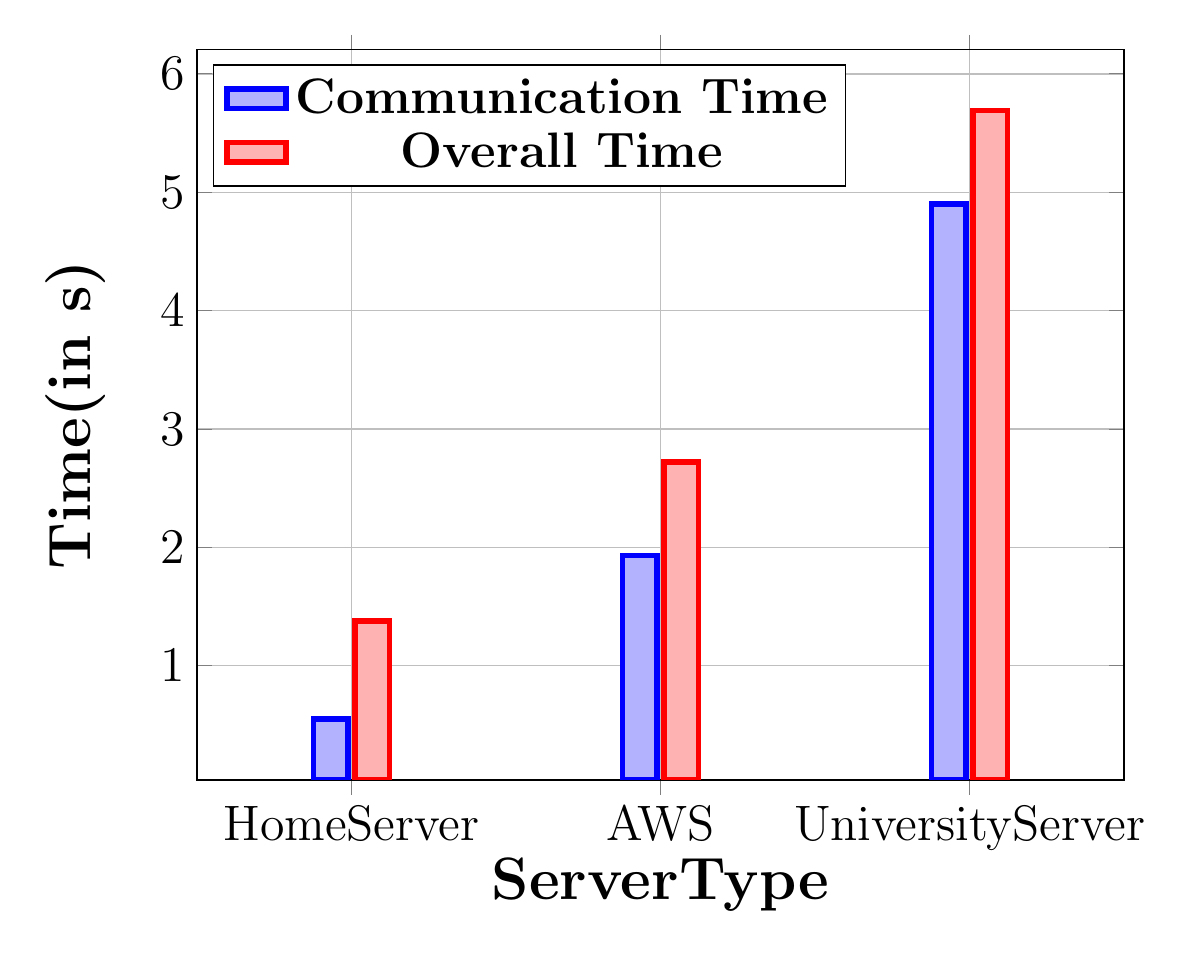}
\caption{Different servers.}
  \label{fig:server}
\end{subfigure}
\caption{Overall time for (a) different layers after which Model B is created using InceptionResNetV2 on ImageNet(299),  (b) spread of overall time for different models, and (c) different server locations using ResNet152 on ImageNet(224). All these experiments were done using the double split configuration.}
\Description{describes the time taken}
\end{figure*}

\begin{table*}
    \centering
    \begin{tabular}{m{3cm} m{1.7cm} m{1cm} m{1cm} m{1cm} m{1cm} m{1cm} m{0.8cm} m{0.8cm} m{0.8cm} m{0.8cm}}
        \toprule
        \multirow{2}{*}{Architecture}& \multirow{2}{*}{\# Parameters} & \multicolumn{3}{c}{Epochs needed for convergence} & \multicolumn{2}{c}{Overall} & \multicolumn{2}{c}{Comm.} & \multicolumn{2}{c}{No Split} \\
        \cmidrule{3-11}
        & & 10 & 100 & 1000 & Double & Single & Double & Single & Mobile & Server\\
        \cmidrule{1-11}
        ResNet50~\cite{he2016deep}          & 27M   & 440 & 30 & 14 & 2.55 & 1.74 & 1.60 & 1.07 & 9.6  & 0.02 \\
        ResNet152~\cite{he2016deep}         & 61M   & 370 & 24 & 7  & 2.72 & 1.95 & 1.93 & 1.03 & NA   & 0.05 \\
        DenseNet121~\cite{huang2017densely}       & 8M    & 489 & 38 & 16 & 2.59 & 1.72 & 1.88 & 0.99 & 11.5 & 0.04 \\
        DenseNet201~\cite{huang2017densely}       & 20M   & 465 & 35 & 14 & 2.67 & 1.82 & 1.87 & 0.93 & NA   & 0.08 \\
        InceptionV3~\cite{szegedy2016rethinking}       & 22.5M & 493 & 40 & 23 & 2.61 & 2.1  & 1.63 & 1.21 & NA   & 0.03 \\
        InceptionResNetV2~\cite{szegedy2016inception} & 54M   & 445 & 48 & 27 & 2.38 & 2.3  & 1.60 & 1.19 & NA   & 0.17 \\
        \bottomrule
    \end{tabular}
    \caption{Average time taken per epoch, in seconds, for six different models using \textit{SplitEasy}.}
    \label{tab:archs}
\end{table*}

\subsection{Datasets}\label{sec:data}
Motivated by the popularity of mobile applications that process images we test \textit{SplitEasy} with two of the most popular image datasets:

\smallskip\noindent\textbf{1) CIFAR 100~\cite{cifar100}.} The CIFAR 100 dataset consists of 60000 32 $\times$ 32 colour images in 100 classes, with 600 images per class. We simulate the modified backpropagation algorithms proposed in Section~\ref{sec:backprob} and train it on CIFAR 100 using  VGG~\cite{vgg} and DenseNet~\cite{densenet}.

\smallskip\noindent\textbf{2) ImageNet~\cite{imagenet_cvpr09}.} The ImageNet dataset is a compilation of 224 $\times$ 224 and 299 $\times$ 299 human-annotated images that consists of 14,197,122 images and 1000 classes. As demonstrated by You~\textit{et al.}~\cite{img_training}, training models on the ImageNet dataset requires days in a setup with constrained hardware. We tested \textit{SplitEasy} with three ImageNet subsets of 10, 100, and 1000 images.

\subsection{Experiments}\label{sec:exps}

First, to examine whether the proposed modification to backpropagation works, we simulate the new training procedure on the university server using VGG and DenseNet on CIFAR 100~\cite{cifar100} dataset.  
As shown in Figure~\ref{fig:proof_of_concept}, the test accuracy of our method is very close to the conventional backpropagation procedure. 

Now that our proof of concept is established, we use sample images from the ImageNet dataset to measure the training timing and communication overhead per image when we want to train state-of-the-art models on an iPhone XR. Considering that a server needs days to train a model on the ImageNet~\cite{img_training}, it is expected to take a lot more time in an iPhone. As depicted in Figure~\ref{fig:implementation}, \textit{SplitEasy} requires four API calls for every training epoch. Since mobile operating systems allow background processes to make only a few API calls per day, \textit{SplitEasy} runs a conventional application in the foreground to avoid being blocked by iOS. 
To avoid API blocking issues during experimentation, we assumed that the users would be taking the pictures with a varying frequency and train as and when the new images are generated. Therefore, we decided to run our experiments with a single image and average the time for ten such images to get an estimate about the time consumed to run a single image through the model splits distributed between the client and the server.

We examine two split learning cases, one when the models are split only once (\textit{single split}) and one when they are split twice and the mobile device is responsible for the first and the last part (\textit{double split}). We also compare our results to a no split setting in which the training happens entirely either on the mobile device or the server. Notably, due to hardware constraints, the employed mobile device is able to train only two out of the six models we examined (not marked with NA in Table~\ref{tab:archs}).

\smallskip\noindent\textbf{Finding the Split Location.} In both single and double split setup, choosing where to split is a crucial design decision. To understand this, we do a case study with the InceptionResNetV2 architecture, which has 164 layers in total. The first seven layers of the network and the output shapes are given in Table~\ref{tab:layers}. In Figure~\ref{fig:cutoff}, we show how the communication time and overall time changes with respect to changes in the split location. Note that, as shown in Table~\ref{tab:layers}, the FirstConv refers to the first hidden layer, FirstMaxPool refers to the fourth hidden layer, whereas the SecondMaxPool refers to the seventh hidden layer in the network. It is clear that the overall time depends heavily on where we make the cut and its dominated by the communication time. This is because the size of the data sent depends on the layer where the split is made, and the communication costs increase as the size of the data increases. However, a mobile device with limited computational resources may be incapable of executing a few extra layers. Therefore the best split can be found, either manually or automatically, when considering the available hardware, the architecture of the NN and the network speed. In all the reported experiments we found the best split location manually.

\smallskip\noindent\textbf{Single Split.} In this setup, we split the network into two models such that the input data is accessible only to the first model, and the labels are accessible only to the second model. Table~\ref{tab:archs} shows the time taken to run different NN architectures using this setup.

\smallskip\noindent\textbf{Double Split.} In this setup, we do two splits in the network to divide it into three different models, such that the input data is accessible only to the first model, and the labels are only accessible to the third model.
The runtimes for these experiments are shown in Table~\ref{tab:archs}. When both the first and last model are on the mobile device, it becomes the most secure and private setting as both input data and labels are accessible only to the mobile device. We believe this to be the most ideal setting for private usage. Note that the increased privacy comes at the cost of increased runtime as the communication overhead increases with the number of splits.
In Figure~\ref{fig:scatter}, we show the changes in runtime as the number of parameters increases for three different NN architectures on two datasets. The difference between the least and the highest runtime is around 0.6s despite a three fold increase in the number of parameters.

\smallskip\noindent\textbf{The impact of data on convergence.} In Table~\ref{tab:archs}, we list the average time taken per image per epoch in double split, single split and no split settings for six state-of-the-art neural networks. The overall time is dominated by the communication time needed to exchange the weights. The overall time spent in each epoch increases steadily with an increase in the batchsize of the data. 
Additionally, we list the number of the required epochs for each model to converge for a given batchsize (10, 100 and 1000) of the data. Observe that the required number of epochs decreases as the batchsize increases. Using the information from Table~\ref{tab:archs}, we can compute the approximate time required for training a model for a given batchsize and choose the best settings. Note that the numbers for the required epochs are for models trained from scratch; for pre-trained models, the required number of epochs will be smaller~\cite{pan2010tl}. These statistics are independent of the hardware employed for training.


\begin{table}
    \centering
    \begin{tabular}{llll}
         \toprule
         Layer & Output Shape & Layer & Output Shape \\
         \cmidrule{1-4}
         1) Convolution & $149*149*32$ & 5) Convolution & $73*73*80$ \\
         2) Convolution & $147*147*32$ & 6) Convolution & $71*71*192$ \\
         3) Convolution & $147*147*64$ & 7) MaxPool & $35*35*192$ \\
         4) MaxPool & $73*73*64$ \\
         \bottomrule
    \end{tabular}
    \caption{First 7 layers and output shape of InceptionResNetV2~\cite{szegedy2016inception}}
    \label{tab:layers}
\end{table}

\smallskip\noindent\textbf{Choosing the appropriate server.} The location of the server, and most importantly the connection to it, can also influence the overall runtime. Having the server in the same network as the mobile device will give faster runtimes, but this is not always the case. It is common to use cloud platforms like Amazon Web Services (AWS) or Google Cloud Platform to host the servers.
To see how the performance varies with change in server types, we conduct experiments on three different server configurations: (i) A desktop server on the same network as the mobile device, (ii) An AWS instance with a direct TCP connection to the mobile device (iii) A University server accessible through tunneling services like ngrok (https://ngrok.com).
The results for this experiment are shown in Figure~\ref{fig:server}. It is clear that the time spent in communication is the biggest factor for varying runtimes. When the server and the mobile device are located in the same network, the split network performs the best, whereas when the server and the mobile device are connected through multiple routing channels like in the case of ngrok, it performs the worst.


\section{Related Work}
Split learning~\cite{gupta2018distributed, vepakomma2018split} was first introduced with a particular emphasis on the collaborative yet privacy-aware training of health data. Gupta and Raskar~\cite{gupta2018distributed} and Vepakomma~\textit{et al.}~\cite{vepakomma2018split} introduce three different configurations for split learning: \textit{(i)} single split, where the data and the first layers are kept on the client, \textit{(ii)} double split, where the first and the last layers (data and labels) are kept on the client, and \textit{(iii)} split learning for vertically partitioned data. These papers demonstrate the effectiveness of split learning in comparison to FL~\cite{FL} and show the reduced number of floating-point operations per second (FLOPS) on the client devices.

However, although these works outline the theory and applications of split learning, they do not emphasize on the implementation challenges. A representative example is automatic differentiation (AD)~\cite{AutoDiff}, a technique used for gradient calculation and backpropagation in standard DL frameworks. AD cannot be used as is in split setting due to the presence of disjoint graphs on the client and server. One of the contributions of this work is a solution that creates ad-hoc labels for every split so that the backpropagation carried in different splits mimics the one on a single machine. We provide more details about this technique in Section~\ref{sec:splitLearning}.

Singh~\textit{et al.}~\cite{singh2019detailed}, motivated by the fact that both split and federated learning have emerged as popular techniques for collaborative, privacy-aware training of the DL models~\cite{kairouz2019advances}, compare their overall communication requirements. Thapa~\textit{et al.}~\cite{thapa2020splitfed} unite the two approaches, eliminate their inherent drawbacks, and analyze the change in communication efficiency. Moreover, Gao~\textit{et al.}~\cite{gao2020end} evaluate split learning for Internet-of-Things (IoT) applications and record the time and communication overhead for the split learning setting. However, similarly to other works that analyze the communication cost, their experiments are run in a single Python-based framework. As a result, the authors do not include the costs that arise due to the graph discontinuity in mobile frameworks.

Moreover, Abuadbba~\textit{et al.}~\cite{abuadbba2020can} evaluate the applicability of split learning for 1-dimensional CNN applications, especially with ECG signals as inputs to the model. Similarly, Poirot~\textit{et al.}~\cite{poirot2019split} explore split learning for health care applications while Kairouz~\textit{et al.}~\cite{kairouz2019advances} suggest exploring the parallel server-client architectures by building upon the ideas presented in~\cite{DecoupledNN} and~\cite{FeaturesReplay}. Another variation of split learning proposed by Praneeth ~\textit{et al.}~\cite{nopeek} attempts to reduce the potential leakage via communicated activations by reducing their distance correlation with the raw data and maintaining good model performance. Lastly, Matsubara and Levorato~\cite{matsubara2020neural} propose a framework for object detection on edge devices using split learning. Their framework uses knowledge distillation to train a smaller network that can run on edge devices. While they are able to reduce the size of the data transmitted, the teacher-student algorithm increases the complexity of the training process and they do not consider the problem of graph discontinuity in their framework.

Motivated by the aforementioned limitations, we develop \textit{SplitEasy} to train DL models on mobile devices. In contrast to related works that discuss the challenges of SL using data from simulated settings, we conduct experiments with mobile devices.



\section{Conclusion and Future Work}
We propose \emph{SplitEasy}, a novel framework designed to offer a quiver to anyone who wants to employ split learning for model training in mobile devices. As demonstrated in our experiments, split learning techniques are a solution for training complex DL models on low-end mobile devices. To the best of our knowledge, this is the first work that discusses the implementation of split learning techniques and reports measurements of the associated computational and communication costs on mobile devices. 
As discussed in Section~\ref{sec:experiments}, we implemented \emph{SplitEasy} using React Native and Tensorflow.js in iOS devices. As React Native can be used for cross-platform development, we plan to extend our work to Android devices in the future. Additionally, as gradient transfers account for a major part of communication cost in \emph{SplitEasy}, we plan to explore the effectiveness of existing compression techniques (quantization, distillation, pruning, etc.) in split learning. Furthermore, as on-device training is one of the major components of a typical FL pipeline, we hope to add FL support in \emph{SplitEasy}, which would allow the community to collaboratively train state of the art models on mobile devices. Lastly, we have only explored the supervised training of the models in the split learning settings. Hence, we plan to explore the possibilities of expanding \emph{SplitEasy} to other domains like unsupervised and reinforcement learning.

\clearpage


\bibliographystyle{ACM-Reference-Format}
\bibliography{sample-base}










\end{document}